%% file: main.tex
\documentclass[sigconf]{acmart}

\usepackage{microtype}
\usepackage{multirow}

\AtBeginDocument{%
  }

\copyrightyear{2026}
\acmYear{2026}
\setcopyright{cc}
\setcctype{by}
\acmConference[MM '26]{Proceedings of the 34th ACM International Conference on Multimedia}{November 10--14, 2026}{Rio de Janeiro, Brazil}
\acmBooktitle{Proceedings of the 34th ACM International Conference on Multimedia (MM '26), November 10--14, 2026, Rio de Janeiro, Brazil}
\acmDOI{10.1145/3767308.3835534}
\acmISBN{979-8-4007-2213-4/2026/11}

\begin{document}

\title{Bridging Language and Spherical Space: Object-Centric Control for Text-to-Panorama Generation}

\author{Derui Li}
\affiliation{%
  \institution{Beijing University of Posts and Telecommunications}
  \city{Beijing}
  \country{China}}
\email{deruili@bupt.edu.cn}

\author{Qian Qiao}
\affiliation{%
  \institution{Beijing University of Posts and Telecommunications}
  \city{Beijing}
  \country{China}}
\email{qqiao@bupt.edu.cn}

\author{Yuhao Sun}
\affiliation{%
  \institution{Beijing University of Posts and Telecommunications}
  \city{Beijing}
  \country{China}}
\email{yuhaosun@bupt.edu.cn}

\author{Wenhao Guo}
\affiliation{%
  \institution{Beijing University of Posts and Telecommunications}
  \city{Beijing}
  \country{China}}
\email{whguo@bupt.edu.cn}

\author{Peng Lu}
\correspondingauthor
\affiliation{%
  \institution{Beijing University of Posts and Telecommunications}
  \city{Beijing}
  \country{China}}
\email{lupeng@bupt.edu.cn}

\renewcommand{\shortauthors}{Derui Li, Qian Qiao, Yuhao Sun, Wenhao Guo, \& Peng Lu}

\begin{abstract}
Panoramic image generation is increasingly important for immersive applications such as virtual reality, augmented reality, and 3D content creation. Unlike perspective images, panoramic images represent a viewer-centered $360^\circ$ surrounding space, where directional expressions such as left, right, front, and behind play a central role in spatial understanding. However, existing text-to-panorama methods largely rely on implicit spatial reasoning and often fail to faithfully ground object-level directional descriptions in spherical panoramic scenes. A straightforward alternative is to introduce explicit layouts, but requiring manually specified spatial conditions reduces the flexibility of language-based interaction and does not directly resolve the misalignment between egocentric directional language and panoramic image space. To address this issue, we propose \textbf{PanoCtrl}, an object-centric framework for controllable text-to-panorama generation. Our method explicitly bridges natural language and spherical panoramic space by converting textual descriptions into structured object-level spherical conditions and integrating them into the diffusion process. Specifically, we introduce \textbf{PanoParse}, a text-conditioned parser that predicts object semantics and spherical bounding field-of-view (BFoV) parameters, and \textbf{PanoControl}, which injects object-level semantic and spatial guidance into the diffusion transformer through object-aware attention and spatial residual enhancement. To support this task, we construct \textbf{PanoGround}, a dataset with object-level spherical annotations and diverse directional descriptions for controllable panoramic generation. Extensive experiments demonstrate that \textbf{PanoCtrl} achieves state-of-the-art performance in both spatial alignment and image quality.
\end{abstract}



\begin{teaserfigure}
    \centering
    \includegraphics[width=\textwidth]{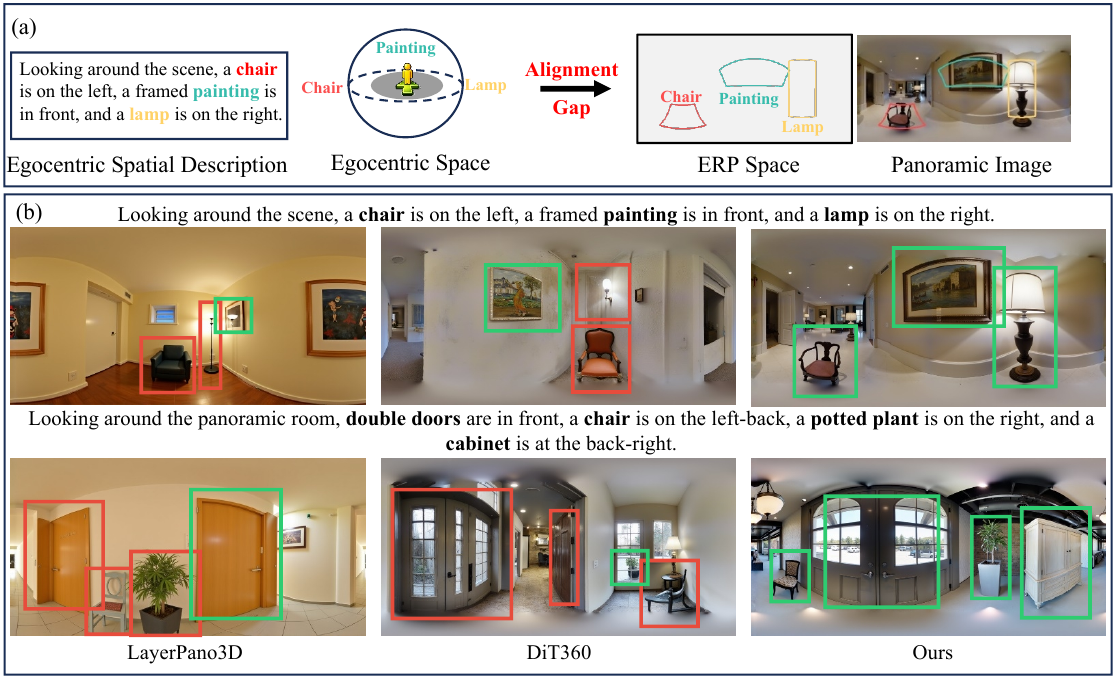}
    \caption{
    Motivation examples for controllable text-to-panorama generation.
    \textbf{(a)} Illustration of the alignment gap between egocentric directional space and ERP space.
    \textbf{(b)} Qualitative comparison with representative methods, including LayerPano3D~\cite{yang2025layerpano3d} and DiT360~\cite{feng2025dit360}, under directional descriptions.
    Red boxes indicate incorrect or spatially inconsistent object placements, while green boxes indicate correctly aligned ones.
    }
    \label{fig:intro}
\end{teaserfigure}

\maketitle

\input{sec/1_intro_v4}
\input{sec/2_relatedwork_v3}

\input{sec/3_method_exp_v6}
\input{sec/5_conclusion_v1}

\begin{acks}
\noindent This work was supported by the BUPT Excellent Ph.D. Students Foundation under Grant No.CX20241087.
\end{acks}

\bibliographystyle{ACM-Reference-Format}
\bibliography{sample-base}

\end{document}

%% file: sec/1_intro_v4.tex
\section{Introduction}

Panoramic image generation has attracted increasing attention in immersive applications such as virtual reality, augmented reality, and 3D content creation~\cite{wang2025conditional, shi2025dreamrelation}. 
Unlike conventional perspective images, panoramas represent a full $360^\circ$ viewer-centered environment, where spatial relationships are defined around the observer rather than on a fixed 2D image plane. 
Therefore, directional expressions such as left, right, front, and behind become fundamental spatial semantics that determine object placement, making accurate directional control critical for panoramic generation.

With the rapid progress of text-to-image diffusion models~\cite{ramesh2021zero,ramesh2022hierarchical,rombach2022high,labs2025flux1kontextflowmatching_flux,esser2024scaling,liu2024playground,guo2026casr}, recent works have extended image generation from perspective images to panoramic scenes~\cite{yang2025layerpano3d,feng2025dit360,zhang2024taming}. 
Although these methods synthesize visually plausible panoramas, they still struggle to ground object-level directional descriptions in complex scenes. As shown in Fig.~\ref{fig:intro}(b), generated objects are often misplaced or spatially inconsistent with the input text, revealing a fundamental gap between language-specified directions and ERP-space representation.

A straightforward solution is to introduce explicit spatial conditions, such as layouts or bounding boxes. However, existing controllable generation methods are mainly designed for perspective images with Euclidean coordinates, and are difficult to directly apply to panoramas due to ERP distortion and spherical wrap-around continuity. Moreover, manually specified layouts reduce the flexibility of language-based interaction. Thus, the key challenge is to achieve language-driven directional control in viewer-centered spherical space.

This leads to two key challenges. First, panoramic generation requires accurately grounding object-level directional descriptions in a $360^\circ$ viewer-centered space, where spherical geometry and ERP representation make spatial alignment significantly more complex than in perspective images. Second, it is challenging to transform natural language into structured spatial conditions and integrate them into the generation process in an end-to-end manner, as decoupled pipelines often lead to inconsistencies between language understanding and image generation, resulting in degraded controllability and spatial misalignment.

To address these challenges, we propose \textbf{PanoCtrl}, an object-centric framework that bridges natural language and spherical panoramic space through structured object-level representations. Specifically, \textbf{PanoParse} converts text descriptions into spherical object conditions, including semantic categories and BFoV (Bounding Field-of-View)~\cite{xu2022pandora} parameters, while \textbf{PanoControl} injects object-aware semantic and spatial guidance into the diffusion transformer through attention and spatial enhancement mechanisms. Furthermore, we construct \textbf{PanoGround}, a dataset with object-level spherical annotations, and establish a benchmark where \textbf{PanoCtrl} achieves state-of-the-art performance in spatial alignment and image quality.

Our contributions are summarized as follows:
\begin{itemize}
    \item We propose \textbf{PanoCtrl}, an object-centric framework for controllable text-to-panorama generation, which explicitly bridges natural language and spherical panoramic space for more accurate directional grounding.
    \item We introduce \textbf{PanoParse} to convert textual descriptions into structured object-level spherical conditions, and \textbf{PanoControl} to inject object-aware semantic and spatial guidance into the diffusion process.
    \item We construct \textbf{PanoGround}, a dataset with object-level spherical annotations for controllable panoramic generation, and establish a benchmark on this task, where \textbf{PanoCtrl} achieves state-of-the-art performance in both spatial alignment and image quality.
\end{itemize}

%% file: sec/2_relatedwork_v3.tex
\section{Related Work}

\subsection{Panoramic Image Generation}

Panoramic images have become increasingly important in immersive applications such as virtual reality, augmented reality, and 3D content creation, as well as spatial perception tasks in embodied intelligence~\cite{wang2025conditional, shi2025dreamrelation, sun2025tactile, sun2025soft, sun2026tacchi}. Existing panoramic image generation methods mainly follow two directions: generating panoramas from multiple perspective views, or modeling panoramic generation more directly in the equirectangular domain.

The first direction synthesizes multiple perspective views and stitches them into a full panorama~\cite{bar2023multidiffusion,li2023panogen,tang2024mvdiffusion++,yu2023long}. These methods leverage the strong priors of conventional perspective generation models, but often suffer from cross-view inconsistencies such as object duplication, structural discontinuities, and weakened global coherence. The second direction introduces panoramic-aware modeling in ERP space or jointly models panoramic and perspective representations~\cite{chen2022text2light,zhang2024taming,ye2024diffpano,wang2024customizing,feng2025dit360,team2025hunyuanworld,yang2025layerpano3d,yang2025matrix3d,worldgen2025ziyangxie,sun2025spherical,park2025spherediff}. Representative efforts improve panorama continuity through dedicated fine-tuning~\cite{wang2024customizing}, incorporate panoramic-aware designs into diffusion models~\cite{ye2024diffpano,feng2025dit360}, or jointly model perspective and panoramic representations to enhance global coherence~\cite{zhang2024taming}. Tuning-free panorama generation has also been explored as a lightweight alternative~\cite{liu2024panofree}. Nevertheless, directly generating ERP panoramas remains challenging because wrap-around continuity and severe polar distortions often reduce geometric fidelity.

Despite these advances, most text-to-panorama methods still rely on global text conditioning and implicit spatial reasoning, limiting object-level directional grounding. Moreover, they overlook the misalignment between egocentric directions and ERP-based representations, which hinders precise spatial control.

\subsection{Controllable Image Generation}

Controllable image generation aims to provide fine-grained control over object placement, layout, and spatial relationships in generated images~\cite{xie2023boxdiff,zhang2023adding,li2023gligen,zheng2023layoutdiffusion}. Existing methods typically introduce explicit structural conditions, such as bounding boxes, segmentation maps, or keypoints, and inject them into diffusion models through conditional attention, feature modulation, or dedicated control modules.

A major line of work focuses on layout-guided generation. Many methods incorporate layout conditions through conditional attention or feature interaction mechanisms~\cite{feng2024ranni,jia2024ssmg,lv2024place,wu2025ifadapter,xue2023freestyle,yang2023reco,zheng2023layoutdiffusion,cheng2023layoutdiffuse,li2023gligen,wang2024instancediffusion}, while others design dedicated layout encoding modules for structured spatial control~\cite{xiang2025instanceassemble,cheng2024hico,yang2023law,zhang2023adding,zhou2024migc}. Autoregressive paradigms have also been explored for spatial planning and generation~\cite{he2025plangen}.

However, most existing methods are designed for perspective images and assume a Euclidean image plane. They cannot be directly applied to panoramas, where spherical geometry, ERP distortion, and wrap-around continuity make directional concepts such as left, right, front, and behind difficult to model. Moreover, they usually require predefined spatial conditions rather than deriving structured controls from natural language. In contrast, our method predicts object-level spherical conditions from text and integrates them into the diffusion process for explicit directional control in panoramic generation.

%% file: sec/3_method_exp_v6.tex
\begin{figure*}[t]
    \centering
    \includegraphics[width=0.9\textwidth]{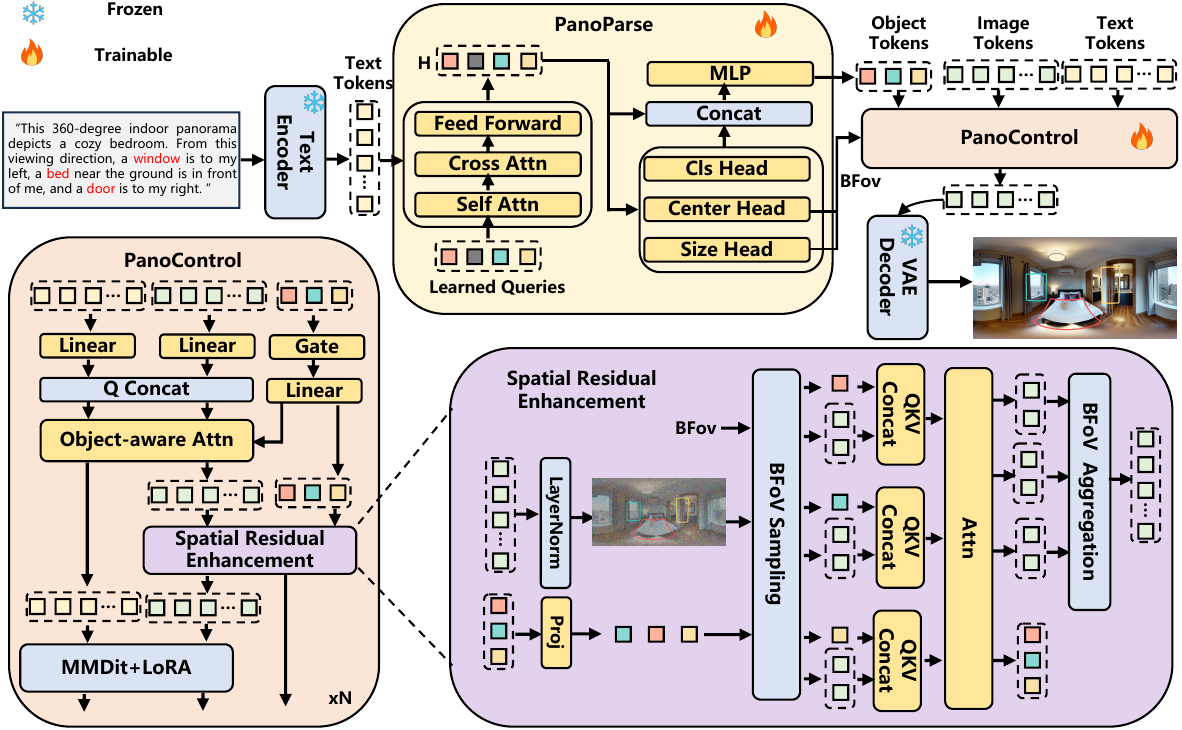}
    \caption{Overall framework of \textbf{PanoCtrl}. \textbf{PanoParse} converts the input prompt into object-level spherical conditions, including semantic categories and BFoV parameters, and \textbf{PanoControl} integrates them into the diffusion transformer through object-aware attention and spatial residual enhancement for controllable panoramic image generation.}
    \label{fig:framework}
\end{figure*}

\section{Method}
\subsection{Overview}

Given a text prompt containing both global scene semantics and object-level directional descriptions, our goal is to generate a panoramic image that is globally coherent while accurately grounding these directions in spherical panoramic space. As shown in Fig.~\ref{fig:framework}, we propose \textbf{PanoCtrl}, a controllable text-to-panorama generation framework consisting of two main modules: \textbf{PanoParse} and \textbf{PanoControl}.

The key challenge is to align egocentric directional language with ERP-based panoramic representation and integrate such structured spatial information into the generation process in an end-to-end manner. To this end, \textbf{PanoParse} first transforms the input prompt into structured \textbf{object-level spherical conditions} by predicting the semantic category and spherical BFoV (Bounding Field-of-View) of each object, thereby converting implicit directional descriptions into explicit object-centric spatial representations. Based on these parsed conditions, \textbf{PanoControl} further encodes object semantics and spherical geometry into unified object control tokens, which are injected into the diffusion transformer through two complementary branches: an \textbf{object-aware attention} branch for semantic binding and a \textbf{spatial residual enhancement} branch for explicit positional guidance on the ERP token grid.

With this design, \textbf{PanoCtrl} explicitly bridges natural language and spherical panoramic space, enabling accurate object-level directional grounding and end-to-end spatial controllability while maintaining high visual quality.

\subsection{PanoParse}

The \textbf{PanoParse} module aims to infer object-level semantics and spherical spatial configurations directly from text. This is challenging because directional expressions in language are implicit, relative to the viewer, and cannot be directly used as explicit control signals in ERP space. Instead of relying on the generative model to implicitly resolve this gap, we formulate text-to-panorama understanding as a set prediction problem in spherical space.

\subsubsection{Text-conditioned Object Decoding}

Let $\mathbf{T}=\{t_j\}_{j=1}^{M}$ denote the text features extracted from the text encoder of the diffusion model. We introduce a fixed set of learnable object queries $\mathbf{Q}=\{q_i\}_{i=1}^{N}$ and feed them into a transformer decoder conditioned on $\mathbf{T}$:
\begin{equation}
\mathbf{H} = \mathrm{PanoParse}_{\mathrm{dec}}(\mathbf{Q}, \mathbf{T}),
\end{equation}
where $\mathbf{H}=\{h_i\}_{i=1}^{N}$ are the decoded object features. Each query attends to the text features and captures a potential object instance described in the prompt. Following DETR-style set prediction~\cite{carion2020end_detr}, we introduce an additional \textit{no-object} class so that the parser can naturally handle varying numbers of objects without predefined ordering.

\subsubsection{Spherical Condition Prediction}

For each decoded object feature $h_i$, the prediction head estimates both semantic and geometric attributes:
\begin{equation}
(c_i, b_i) = \mathrm{PanoParse}_{\mathrm{head}}(h_i), \quad i=1,\dots,N,
\end{equation}
where $c_i$ denotes the semantic category and
$b_i=(\theta_i,\phi_i,\alpha_i,\beta_i)$ denotes the corresponding spherical BFoV, including center longitude $\theta_i$, center latitude $\phi_i$, and horizontal and vertical angular spans $\alpha_i$ and $\beta_i$.

The prediction head contains three branches: a category branch, a center branch, and a size branch. For BFoV center prediction, we adopt a coarse-to-fine formulation by discretizing longitude and latitude into angular bins and then regressing continuous offsets within the selected bins for refinement. Concretely, the center loss consists of a bin classification loss and an offset regression loss:
\begin{equation}
\mathcal{L}_{\text{center}}^{(i)} =
\mathcal{L}_{\text{bin}}^{(i)} + \lambda_{\text{off}} \mathcal{L}_{\text{off}}^{(i)},
\end{equation}
where $\mathcal{L}_{\text{bin}}^{(i)}$ supervises the predicted longitude and latitude bins, and $\mathcal{L}_{\text{off}}^{(i)}$ regresses the corresponding within-bin offsets. This design is particularly important for panoramic generation, where object locations must be defined in spherical space rather than local planar image space. For BFoV size prediction, we directly regress the horizontal and vertical angular spans.

\subsubsection{Parser Supervision}

We train the parser using bipartite matching following DETR~\cite{carion2020end_detr}. After matching predicted object queries with ground-truth annotations, we apply classification and localization losses to the matched pairs:
\begin{equation}
\mathcal{L}_{\text{parse}} =
\sum_{i \in \mathcal{M}} \left(
\mathrm{CE}(c_i, c_i^{*})
+ \lambda_{\text{ctr}} \mathcal{L}_{\text{center}}^{(i)}
+ \lambda_{\text{size}} \left\| (\alpha_i,\beta_i) - (\alpha_i^{*},\beta_i^{*}) \right\|_1
\right),
\label{eq:parse_loss}
\end{equation}
where $\mathcal{M}$ denotes the matched query set, and $c_i^{*}$ and $(\theta_i^{*},\phi_i^{*},\alpha_i^{*},\beta_i^{*})$ denote the corresponding ground-truth annotations. Unmatched queries are supervised using the \textit{no-object} class.

This loss encourages accurate prediction of both object semantics and spherical locations, providing structured supervision for language-to-spherical parsing and forming the basis for subsequent controllable generation.

\subsection{PanoControl}

Given the parsed object features $\{h_i\}_{i=1}^{N}$ and the predicted object-level spherical conditions $\{(c_i,b_i)\}_{i=1}^{N}$, \textbf{PanoControl} transforms them into structured control signals for diffusion-based panoramic generation.

Although diffusion transformers are effective at modeling global scene semantics, they do not explicitly enforce accurate object placement in ERP space. To address this limitation, we introduce a dual-branch control design that combines \textbf{object-aware attention} and \textbf{spatial residual enhancement}. The former improves semantic binding between object conditions and generated content, while the latter provides explicit region-level positional guidance on the ERP token grid. Their combination enables both semantic faithfulness and accurate spatial grounding.

\subsubsection{Object Condition Construction}

We first convert each parsed object into a unified control token that jointly encodes object semantics and spherical geometry:
\begin{equation}
o_i = \mathrm{MLP}\big([h_i,\ e_i^{\text{cls}},\ p_i^{\text{pos}}]\big),
\end{equation}
where $h_i$ is the decoded object feature from \textbf{PanoParse}, $e_i^{\text{cls}}$ is the embedding of the predicted object category, and $p_i^{\text{pos}}$ encodes the spherical BFoV parameters. In practice, $p_i^{\text{pos}}$ is constructed from the predicted BFoV center and angular spans, providing an explicit positional representation in spherical space.

Only valid object tokens, i.e., those not assigned to the \textit{no-object} class, are retained for condition injection. These object tokens serve as the unified interface between parsing and generation, and are subsequently injected into the diffusion transformer for both semantic and spatial control. In our experiments, the default number of object queries is set to 8, which provides a good trade-off between representation capacity and generation stability.

\subsubsection{Object-aware Attention Injection}

To introduce object-level semantic control into the denoising process, we inject the object tokens into the diffusion transformer through gated attention. For each object token $o_i$, we first compute a learnable gating factor:
\begin{equation}
\tilde{o}_i = \sigma(\mathrm{MLP}(o_i)) \cdot o_i,
\end{equation}
and denote the gated object token set as $\tilde{O}=\{\tilde{o}_i\}_{i=1}^{N}$.

Let $T$ and $I$ denote the text tokens and image tokens in an MMDiT block~\cite{esser2024scaling,labs2025flux1kontextflowmatching_flux}. We inject the object conditions into both streams by:
\begin{equation}
\hat{T} = T + \mathrm{Attn}(Q=T,\ K=\tilde{O},\ V=\tilde{O}),
\end{equation}
\begin{equation}
\hat{I} = I + \mathrm{Attn}(Q=I,\ K=\tilde{O},\ V=\tilde{O}),
\end{equation}
where $\hat{T}$ and $\hat{I}$ are the conditioned text and image tokens, respectively.

This branch improves the consistency between object appearance and textual descriptions by explicitly introducing object-level semantics into the denoising process. However, attention-based conditioning mainly operates implicitly through global token interaction and is therefore insufficient for precise positional control in ERP space.

\subsubsection{Spatial Residual Enhancement}

To provide explicit region-level guidance in ERP space, we introduce a spatial residual enhancement branch that enhances the image token grid according to the predicted BFoVs. As shown in Fig.~\ref{fig:framework}, this branch consists of three steps: \textbf{BFoV Sampling}, local feature interaction, and \textbf{BFoV Aggregation}.

Given object tokens $\{o_i\}_{i=1}^{N}$ and their predicted BFoVs $\{b_i\}_{i=1}^{N}$, we first project each object token into the image-token space:
\begin{equation}
r_i = \mathrm{Proj}(o_i).
\end{equation}
Here, each $b_i$ is represented by differentiable box parameters, which define the object region in spherical space and allow the spatial enhancement branch to be optimized end-to-end, similar to prior differentiable box-based region control methods~\cite{lu2020weakly_crop}.

In the \textbf{BFoV Sampling} step, each $b_i$ is used to sample the corresponding local region from the ERP image token grid, producing BFoV-aligned local features $\tilde{I}_i$. These sampled local features are then concatenated with $r_i$ and passed through QKV transformation and attention to obtain enhanced local representations:
\begin{equation}
A_i = \mathrm{Attn}\bigl(\mathrm{QKV}([\tilde{I}_i; r_i])\bigr),
\end{equation}
where $[\cdot;\cdot]$ denotes concatenation.

In the \textbf{BFoV Aggregation} step, the enhanced local features are written back to the global ERP token grid via residual fusion:
\begin{equation}
\bar{I} = \hat{I} + \alpha \sum_{i=1}^{N} \mathrm{Agg}(A_i),
\label{eq:spatial_residual}
\end{equation}
where $\mathrm{Agg}(\cdot)$ maps the enhanced BFoV features to their corresponding ERP locations, and $\alpha$ controls the enhancement strength. When multiple objects overlap, their contributions are accumulated on the global token grid.

Compared with object-aware attention, which affects generation mainly through global token interaction, this branch provides more direct region-level guidance by explicitly sampling and aggregating features within BFoV-specified regions. This design is especially beneficial for reducing the mismatch between egocentric spatial descriptions and ERP-space panoramic generation.

Finally, the conditioned text tokens $\hat{T}$ and enhanced image tokens $\bar{I}$ are fed into subsequent diffusion transformer blocks for controllable panoramic image synthesis.

\subsection{Training Objectives}

Our framework is trained end-to-end by jointly optimizing the parsing objective and the diffusion objective.

\subsubsection{Object Parsing Objective}

The \textbf{PanoParse} module is supervised using object-level spherical annotations from the training set. Following the set-prediction paradigm, we optimize the parsing loss $\mathcal{L}_{\text{parse}}$ in Eq.~\ref{eq:parse_loss}, which encourages accurate prediction of both object semantics and spherical locations and provides explicit structural supervision for text-to-spherical parsing.

\subsubsection{Diffusion Objective}

Given the global prompt and the object-level control signals produced by \textbf{PanoControl}, we train the diffusion transformer using a standard global denoising loss together with an additional object-region loss. The global diffusion loss is:
\begin{equation}
\mathcal{L}_{\text{global}} = \mathbb{E}_{x_0,\epsilon,t}\left[
\|\epsilon - \epsilon_\theta(x_t, t, \mathcal{C})\|_2^2
\right],
\label{eq:global_loss}
\end{equation}
where $x_t$ is the noisy latent at timestep $t$, $\epsilon$ is the sampled Gaussian noise, and $\mathcal{C}$ denotes all conditioning inputs, including the prompt and the object-level control signals.

To further strengthen local object consistency, we additionally apply a masked loss over regions covered by the ground-truth BFoVs:
\begin{equation}
\mathcal{L}_{\text{obj}} = \mathbb{E}_{x_0,\epsilon,t}\left[
\|M_{\text{gt}} \odot (\epsilon - \epsilon_\theta(x_t, t, \mathcal{C}))\|_2^2
\right],
\label{eq:obj_loss}
\end{equation}
where $M_{\text{gt}}$ denotes the union of ground-truth BFoV regions. The final diffusion objective is:
\begin{equation}
\mathcal{L}_{\text{diff}} = \mathcal{L}_{\text{global}} + \lambda_{\text{obj}} \mathcal{L}_{\text{obj}}.
\label{eq:diff_loss}
\end{equation}

This object-region loss encourages the model to allocate more denoising capacity to BFoV-specified regions, thereby improving local object fidelity and spatial consistency.

\subsubsection{Overall Objective}

The full model is trained by jointly optimizing the parsing and diffusion objectives:
\begin{equation}
\mathcal{L} = \mathcal{L}_{\text{parse}} + \lambda \mathcal{L}_{\text{diff}},
\label{eq:overall_loss}
\end{equation}
where $\lambda$ balances structured parsing and controllable image generation.

\subsection{PanoGround Dataset}
\label{sec:dataset}

To support controllable panoramic generation with object-level spherical supervision, we construct a new dataset named \textbf{PanoGround}. The dataset is built from publicly available panoramic sources, including LayerPano3D-PanoData~\cite{yang2025layerpano3d}, Matterport3D~\cite{chang2017matterport3d}, and PANDORA~\cite{xu2022pandora}. In total, PanoGround contains 12,688 ERP panoramic images and 37,980 annotated records covering 106 object categories.

Each record consists of a panoramic image, a scene-level description, an object set, and spherical BFoV annotations. We adopt a semi-automatic multi-stage annotation process. Specifically, we first perform scene understanding using a vision-language model~\cite{liu2023visual}, then apply an open-vocabulary detector~\cite{liu2024grounding} to obtain candidate object regions and semantic labels, and finally refine the generated caption variants with a large language model~\cite{wang2024qwen2}, followed by human verification.

Compared with existing panoramic datasets, PanoGround explicitly provides object-level spherical annotations together with diversified directional descriptions, making it suitable for both text-to-layout learning and controllable panoramic generation.

\section{Experiments}

\subsection{Experimental Setup}

\subsubsection{Dataset}

We conduct all experiments on the proposed \textbf{PanoGround} dataset. PanoGround contains 12,688 panoramic images and 37,980 annotated records with object-level spherical BFoV annotations, providing supervision for both text-to-layout parsing and controllable panoramic image generation. Following the dataset setting described in Sec.~\ref{sec:dataset}, we split the dataset into 10,820 panoramas for training and 1,868 panoramas for testing, corresponding to 32,385 and 5,595 annotated records, respectively.

\subsubsection{Implementation Details}

Our method is built upon a diffusion transformer with a Flux-style backbone~\cite{labs2025flux1kontextflowmatching_flux}. To enable efficient adaptation, we fine-tune the model using LoRA~\cite{hu2022lora}. All panoramic images are represented in ERP format with an input resolution of $1024 \times 512$. The model is trained for 20k iterations with a batch size of 4 on 8 NVIDIA A6000 GPUs. During training, object-level control tokens are first predicted by \textit{PanoParse} and then injected into the diffusion transformer through \textit{PanoControl}. Unless otherwise specified, the spatial blending coefficient is set to $\alpha=0.7$, and the number of object queries is set to 8.

\subsubsection{Baselines}

We compare our method with several recent text-to-panorama generation approaches, including \textit{PanFusion}~\cite{zhang2024taming}, \textit{SMGD}~\cite{sun2025spherical}, \textit{PAR}~\cite{wang2025conditional}, \textit{WorldGen}~\cite{worldgen2025ziyangxie}, \textit{Matrix-3D}~\cite{yang2025matrix3d}, \textit{LayerPano3D}~\cite{yang2025layerpano3d}, \textit{HunyuanWorld}~\cite{team2025hunyuanworld}, and \textit{DiT360}~\cite{feng2025dit360}. These baselines cover representative panorama generation paradigms, including both ERP-native panoramic generators and large-scale panoramic world generation models. For fair comparison, all methods are evaluated on the same test split using the same text prompts.

\subsubsection{Evaluation Metrics}

We evaluate controllable panoramic image generation from two aspects: 
\textbf{spatial alignment} and \textbf{image quality}. 
For spatial alignment, we adopt three metrics: 
\textbf{Object Presence Rate (OPR)}, 
\textbf{Region-Text Alignment (RTA)}, and 
\textbf{Spherical Localization Error (SLE)}. 
Note that directional control allows multiple plausible layouts within the same sector; therefore, OPR and RTA measure sector-level correctness rather than exact position matching.

\textbf{Object Presence Rate (OPR).}
For each generated ERP panorama, we project it into perspective views corresponding to predefined egocentric directions and use a vision-language model~\cite{wang2024qwen2} to determine whether the target object appears in the required directional sector. 
We report the average success rate over all annotated object-direction pairs:
\begin{equation}
\mathrm{OPR} = \frac{1}{N}\sum_{i=1}^{N}\mathbb{I}(a_i = 1),
\end{equation}
where $a_i \in \{0,1\}$ indicates whether the object is correctly generated in the target sector.

\textbf{Region-Text Alignment (RTA).}
For each object-direction pair, we render the corresponding directional view and compute its similarity with the object-level text phrase using a vision-language model~\cite{radford2021learning}:
\begin{equation}
\mathrm{RTA} = \frac{1}{N}\sum_{i=1}^{N}
\mathrm{Sim}(v_i^{\mathrm{view}}, t_i^{\mathrm{obj}}),
\end{equation}
where $v_i^{\mathrm{view}}$ denotes the rendered view and $t_i^{\mathrm{obj}}$ denotes the object-level phrase.

\textbf{Spherical Localization Error (SLE).}
SLE provides a complementary localization diagnostic by measuring the angular distance between the generated object center $(\theta_i,\phi_i)$ and the target BFoV center $(\theta_i^*,\phi_i^*)$:
\begin{equation}
\mathrm{SLE} = \frac{1}{|\mathcal{M}|}\sum_{i\in\mathcal{M}}
\arccos(
\sin\phi_i\sin\phi_i^*
+
\cos\phi_i\cos\phi_i^*
\cos(\theta_i-\theta_i^*)
),
\end{equation}
reported in degrees. 
Unlike OPR/RTA, SLE provides a finer-grained localization measurement with respect to the target sector center. Therefore, positional variations within a valid sector may increase SLE but do not affect sector-level correctness.

\textbf{Image Quality.}
We further evaluate visual quality using \textbf{FID}~\cite{heusel2017gans}, 
\textbf{FAED}~\cite{oh2022bips}, 
\textbf{IS}~\cite{salimans2016improved}, and 
\textbf{CLIP Score (CS)}~\cite{radford2021learning}. 
FID measures realism by comparing feature distributions, FAED evaluates panoramic structural quality, IS measures diversity and object discriminability, and CLIP Score measures semantic alignment.

\begin{table}[t]
\centering
\small
\setlength{\tabcolsep}{3pt}
\caption{Quantitative comparison on PanoGround.}
\label{tab:main_results}
\resizebox{\columnwidth}{!}{
\begin{tabular}{lccccccc}
\toprule
& \multicolumn{3}{c}{Spatial Alignment} 
& \multicolumn{4}{c}{Image Quality} \\
\cmidrule(lr){2-4} \cmidrule(lr){5-8}
Method 
& OPR$\uparrow$ & RTA$\uparrow$ & SLE$\downarrow$ 
& FID$\downarrow$ & FAED$\downarrow$ & IS$\uparrow$ & CS$\uparrow$ \\
\midrule
PanFusion    & 76.61 & 19.87 & 52.30 & 86.31 & 9.46 & 2.59 & 23.21 \\
SMGD         & 80.08 & 22.63 & 62.27 & 77.48 & 5.10 & 2.97 & 25.24 \\
PAR          & 73.94 & 23.02 & 58.88 & 70.40 & 6.86 & 3.19 & 27.86 \\
WorldGen     & 77.07 & 23.01 & 48.96 & 60.71 & 4.78 & 3.31 & 29.42 \\
Matrix-3D    & 78.01 & 24.19 & 53.76 & 51.78 & 3.05 & 3.16 & 30.53 \\
LayerPano3D  & 78.66 & 23.58 & 50.58 & 56.22 & 3.89 & 3.05 & 27.64 \\
HunyuanWorld & 85.33 & 26.27 & 53.12 & 50.74 & 2.81 & 3.22 & 28.58 \\
DiT360       & 83.33 & 27.68 & 47.63 & 47.06 & 2.82 & 3.37 & 29.96 \\
\midrule
\textbf{PanoCtrl} & \textbf{98.59} & \textbf{36.91} & \textbf{25.34} & \textbf{46.86} & \textbf{2.77} & 3.35 & \textbf{32.54} \\
\bottomrule
\end{tabular}}
\end{table}

\begin{figure*}[t]
\centering
\includegraphics[width=\linewidth]{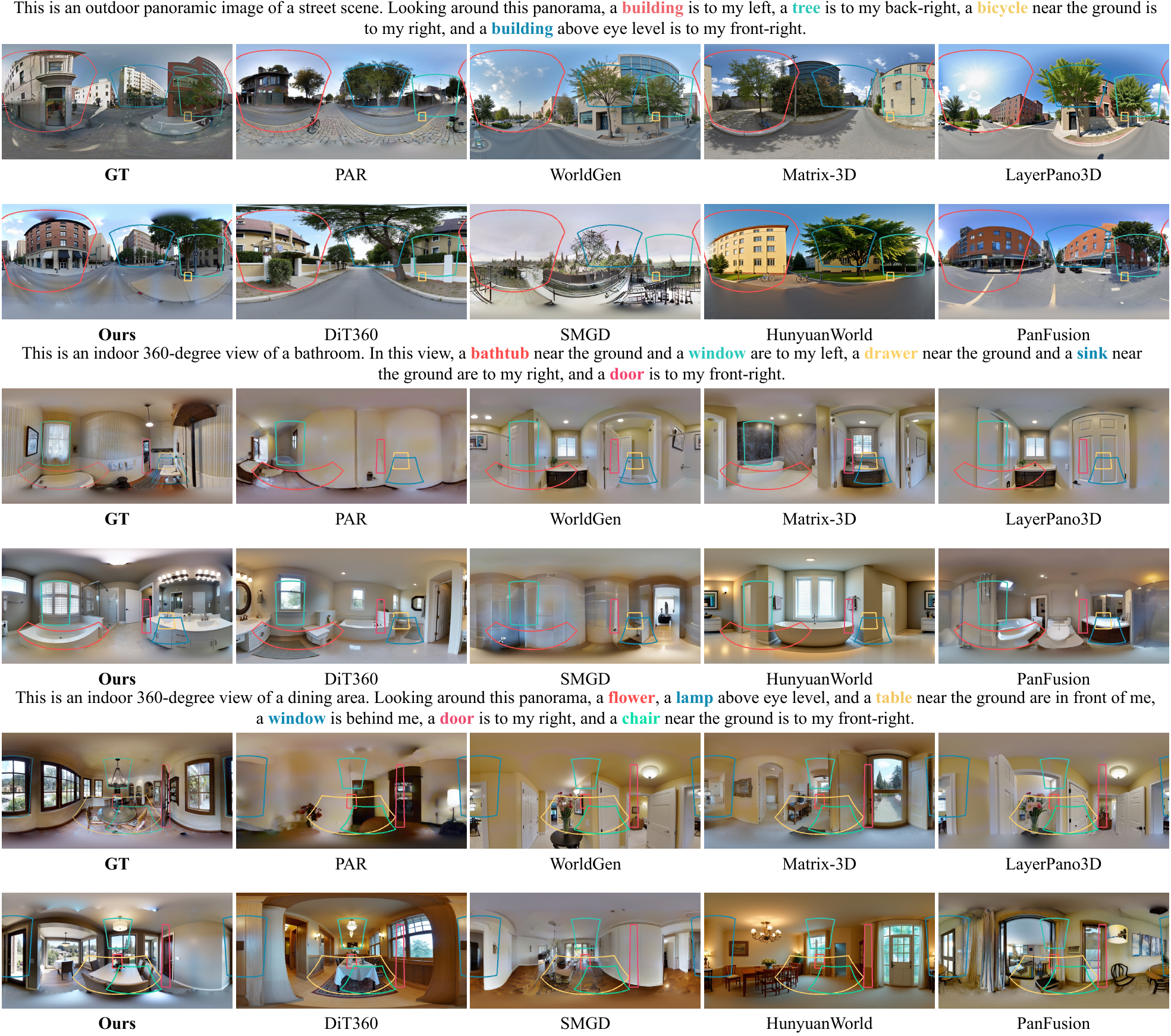}
\caption{Qualitative comparison on challenging panoramic scenes with multi-object directional descriptions. Captions are derived from the ground-truth panoramas, and colored phrases correspond to the colored boxes in the images. Compared with prior methods, \textbf{Ours} achieves more accurate directional grounding and more coherent layouts.}
\label{fig:qualitative}
\end{figure*}

\subsection{Comparison with State-of-the-Art}
\subsubsection{Quantitative Results}

We compare our method with state-of-the-art text-to-panorama generation approaches on the \textbf{PanoGround} benchmark. As shown in Table~\ref{tab:main_results}, \textbf{PanoCtrl} achieves the best overall performance, especially in spatial alignment.

For spatial alignment, \textbf{PanoCtrl} achieves the highest OPR of \textbf{98.59}, surpassing HunyuanWorld by \textbf{13.26}, and the best RTA of \textbf{36.91}, exceeding DiT360 by \textbf{9.23}. More importantly, it reduces SLE to \textbf{25.34}, compared with \textbf{47.63} achieved by the strongest competing method. These results indicate substantially more accurate grounding of directional language in spherical panoramic space.

In terms of image quality, \textbf{PanoCtrl} also delivers strong performance. It achieves the best FID of \textbf{46.86}, the best FAED of \textbf{2.77}, and the highest CLIP score of \textbf{32.54} among all methods. Although its IS reaches \textbf{3.35}, which is slightly lower than the \textbf{3.37} of DiT360, the gap is marginal. Overall, these results show that explicit object-level spherical control improves spatial accuracy without sacrificing perceptual realism or semantic consistency.

\subsubsection{Qualitative Results}

Figure~\ref{fig:qualitative} presents qualitative comparisons with representative baselines. Each example contains multiple objects associated with explicit directional descriptions under spherical panoramic geometry, and ground-truth BFoVs are overlaid for reference.

Existing methods often misplace, omit, or ambiguously generate objects specified in particular directions, especially in complex multi-object scenes. They also tend to produce unstable layouts when multiple directional constraints must be satisfied jointly. In contrast, our method places target objects more accurately within the designated BFoVs, preserves clearer spatial separation, and maintains a more coherent overall scene structure. Moreover, while several baselines exhibit distortions or degraded textures under explicit spatial control, our results remain more realistic and visually consistent. These comparisons further verify that \textbf{PanoCtrl} effectively balances spatial controllability and image quality.

\begin{table}[t]
\centering
\footnotesize
\setlength{\tabcolsep}{2.8pt}
\caption{Ablation study of the main components in \textbf{PanoCtrl}.}
\label{tab:ablation_comp}
\resizebox{0.48\textwidth}{!}{
\begin{tabular}{lcccccccc}
\toprule
Model & Parse & Attn & Spa & OPR$\uparrow$ & RTA$\uparrow$ & SLE$\downarrow$ & FID$\downarrow$ & CS$\uparrow$ \\
\midrule
Baseline &  &  &  & 82.47 & 24.35 & 44.82 & 50.94 & 27.83 \\
\midrule
I   & \checkmark &  &  & 88.94 & 27.12 & 42.63 & 49.85 & 29.41 \\
II  & \checkmark &  & \checkmark & 91.32 & 29.76 & 38.18 & 50.82 & 28.58 \\
III & \checkmark & \checkmark &  & 93.87 & 33.41 & 30.12 & 47.96 & 31.67 \\
Ours & \checkmark & \checkmark & \checkmark & \textbf{98.59} & \textbf{36.91} & \textbf{25.34} & \textbf{46.86} & \textbf{32.54} \\
\bottomrule
\end{tabular}}
\end{table}

\begin{table}[t]
\centering
\small
\setlength{\tabcolsep}{4pt}
\caption{Ablation study of the training objectives.}
\label{tab:ablation_loss}
\begin{tabular}{lccccc}
\toprule
Setting & OPR$\uparrow$ & RTA$\uparrow$ & SLE$\downarrow$ & FID$\downarrow$ & CS$\uparrow$ \\
\midrule
w/o $\mathcal{L}_{\text{parse}}$ & 90.26 & 32.84 & 31.92 & 47.28 & 31.42 \\
w/o $\mathcal{L}_{\text{obj}}$   & 96.85 & 35.96 & 27.43 & 47.02 & 32.06 \\
Ours & \textbf{98.59} & \textbf{36.91} & \textbf{25.34} & \textbf{46.86} & \textbf{32.54} \\
\bottomrule
\end{tabular}
\end{table}

\subsection{Ablation Study}

We conduct ablation studies to analyze the contribution of each component and design choice in our framework. In particular, we study three key parts of our method: \textit{PanoParse}, the \textit{object-aware attention} branch in \textit{PanoControl}, and the \textit{spatial residual enhancement} branch in \textit{PanoControl}. We further examine the effects of the training objectives, the spatial blending coefficient, and the number of object queries.

\subsubsection{Main Components}

Table~\ref{tab:ablation_comp} reports the contribution of the main components in \textbf{PanoCtrl}. Starting from the baseline, introducing \textit{PanoParse} already brings clear gains in OPR and RTA, indicating that explicit parsing of object-level spherical conditions provides effective structural guidance for controllable generation. Further adding the spatial residual enhancement branch improves geometric accuracy and significantly reduces SLE, although image quality is slightly affected when semantic interaction is absent. By contrast, introducing object-aware attention leads to consistent improvements in both alignment and perceptual quality, suggesting that semantic interaction is important for faithfully rendering object attributes under spatial constraints. Combining all components yields the best results across all metrics, demonstrating that explicit spherical parsing, semantic interaction, and spatial enhancement are complementary and jointly essential for reducing the alignment gap in controllable panoramic generation.

\subsubsection{Training Objectives}

Table~\ref{tab:ablation_loss} evaluates the contribution of different training objectives. Removing the parsing loss $\mathcal{L}_{\text{parse}}$ causes noticeable degradation in OPR, RTA, and SLE, indicating that accurate structural supervision is essential for reliable controllability. Removing the object-level loss $\mathcal{L}_{\text{obj}}$ also leads to clear performance drops, especially in local alignment and semantic consistency. The full model achieves the best results, confirming that both objectives are necessary for jointly optimizing layout prediction and controllable panoramic generation.

\subsubsection{Spatial Blending Coefficient}

Table~\ref{tab:ablation_alpha} studies the impact of the spatial blending coefficient $\alpha$. When $\alpha$ is too small, the spatial signal is insufficient, resulting in weaker object alignment. Increasing $\alpha$ steadily improves spatial controllability, and the best overall performance is achieved at $\alpha=0.7$. However, setting $\alpha$ too large slightly hurts both alignment and quality, suggesting that overly strong spatial constraints may interfere with global generation consistency. This observation reflects a trade-off between precise spatial control and holistic image coherence.

\subsubsection{Number of Object Queries}

Table~\ref{tab:ablation_query} analyzes the influence of the number of object queries. Using too few queries limits the model’s ability to represent multiple objects, leading to inferior alignment performance. Increasing the number of queries consistently improves both controllability and generation quality. The best spatial performance is obtained with 8 queries, while 16 queries bring only marginal gains in FID and CS without further improving alignment. Therefore, we adopt 8 queries as a good balance between representation capacity and generation stability.
\begin{table}[t]
\centering
\small
\setlength{\tabcolsep}{4pt}
\caption{Ablation study of the spatial blending coefficient $\alpha$.}
\label{tab:ablation_alpha}
\begin{tabular}{cccccc}
\toprule
$\alpha$ & OPR$\uparrow$ & RTA$\uparrow$ & SLE$\downarrow$ & FID$\downarrow$ & CS$\uparrow$ \\
\midrule
0.3 & 95.12 & 34.04 & 33.82 & 47.91 & 31.66 \\
0.5 & 96.67 & 34.78 & 29.96 & 47.08 & 31.84 \\
0.7 & \textbf{98.59} & \textbf{36.91} & \textbf{25.34} & \textbf{46.86} & \textbf{32.54} \\
0.9 & 96.88 & 35.12 & 27.41 & 47.52 & 31.97 \\
\bottomrule
\end{tabular}
\end{table}

\begin{table}[t]
\centering
\small
\setlength{\tabcolsep}{4pt}
\caption{Ablation study of the number of object queries.}
\label{tab:ablation_query}
\begin{tabular}{cccccc}
\toprule
Queries & OPR$\uparrow$ & RTA$\uparrow$ & SLE$\downarrow$ & FID$\downarrow$ & CS$\uparrow$ \\
\midrule
2  & 94.12 & 33.21 & 37.84 & 47.86 & 31.02 \\
4  & 96.43 & 34.98 & 31.62 & 47.22 & 31.74 \\
8  & \textbf{98.59} & \textbf{36.91} & \textbf{25.34} & 46.86 & 32.54 \\
16 & 98.48 & 36.02 & 27.91 & \textbf{46.65} & \textbf{32.81} \\
\bottomrule
\end{tabular}
\end{table}

%% file: sec/5_conclusion_v1.tex
\section{Conclusion}

In this paper, we present \textbf{PanoCtrl}, an object-centric framework for controllable text-to-panorama generation. Our method explicitly bridges natural language and spherical panoramic space by converting textual descriptions into structured object-level spherical conditions and integrating them into the diffusion process. We further construct \textbf{PanoGround}, a dataset with object-level spherical annotations and directional descriptions for this task. Extensive experiments demonstrate that \textbf{PanoCtrl} achieves state-of-the-art performance in both spatial alignment and image quality, highlighting the effectiveness of explicit language-to-spherical modeling for controllable panoramic generation.